%% file: main.tex
\documentclass[10pt,twocolumn,letterpaper]{article}
\usepackage{breqn}
\usepackage[ruled,vlined]{algorithm2e}
\DeclareMathOperator*{\argmax}{argmax}
\usepackage{comment}
\usepackage{float}
\usepackage[accsupp]{axessibility} 

\usepackage{cvpr}              

\input{preamble}

\definecolor{cvprblue}{rgb}{0.21,0.49,0.74}
\usepackage[pagebackref,breaklinks,colorlinks,citecolor=cvprblue]{hyperref}

\def\paperID{49} 
\def\confName{CVPR}
\def\confYear{2024}

\title{Segmentation-Free Guidance for Text-to-Image Diffusion Models}

\author{Kambiz Azarian, Debasmit Das, Qiqi Hou, Fatih Porikli\\
Qualcomm AI Research\thanks{\noindent Qualcomm AI Research is an initiative of Qualcomm Technologies, Inc.}\\
{\tt\small \{kambiza, debadas, qhou, fporikli\}@qti.qualcomm.com}
}

\makeatletter
\g@addto@macro\@maketitle{
\begin{center}
  \begin{tabular}{@{}cccccc@{}}
    \textbf{Classifier-free}  \!\! & \textbf{Segmentation-free}  \!\! &
    \textbf{Classifier-free}  \!\! & \textbf{Segmentation-free}  \!\! &
    \textbf{Classifier-free}  \!\! & \textbf{Segmentation-free} \\
    
    \includegraphics[width=0.15\linewidth]{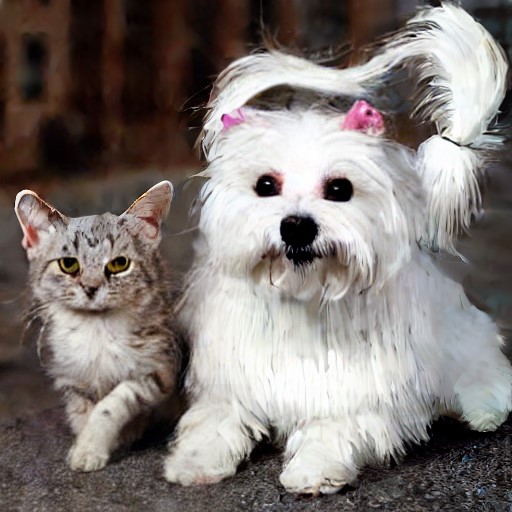} \hspace{-3mm} &
    \includegraphics[width=0.15\linewidth]{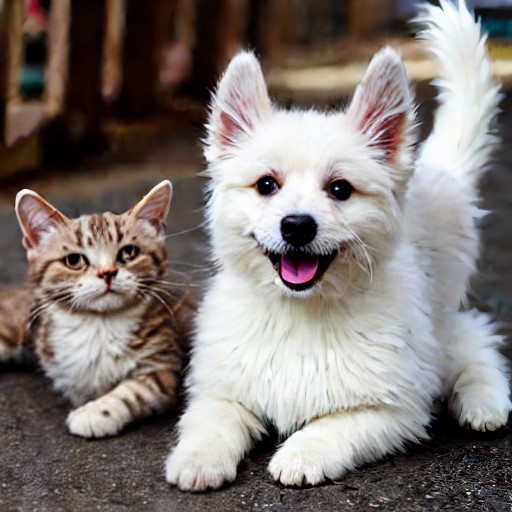}  \!\! &
    \includegraphics[width=0.15\linewidth]{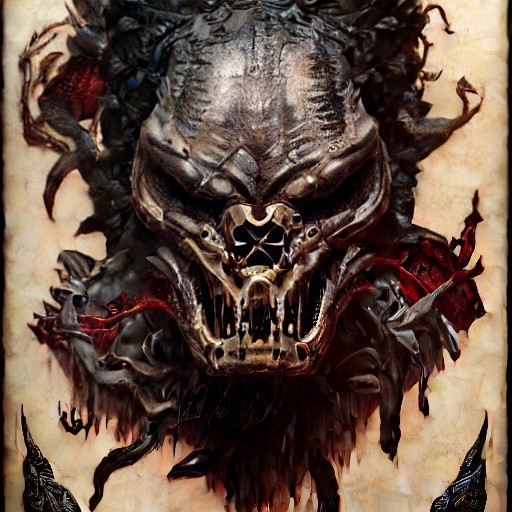}  \hspace{-3mm} &
    \includegraphics[width=0.15\linewidth]{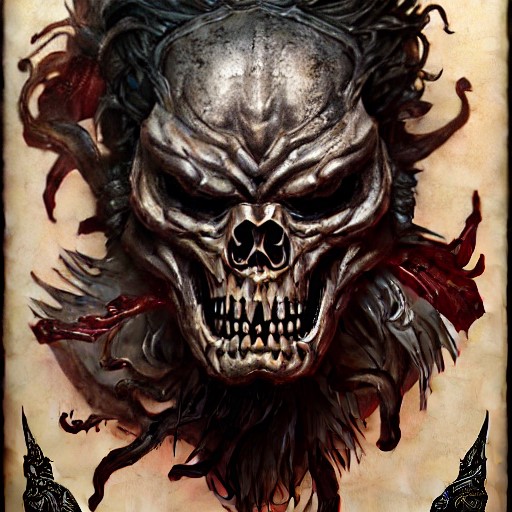}  \!\! &
    \includegraphics[width=0.15\linewidth]{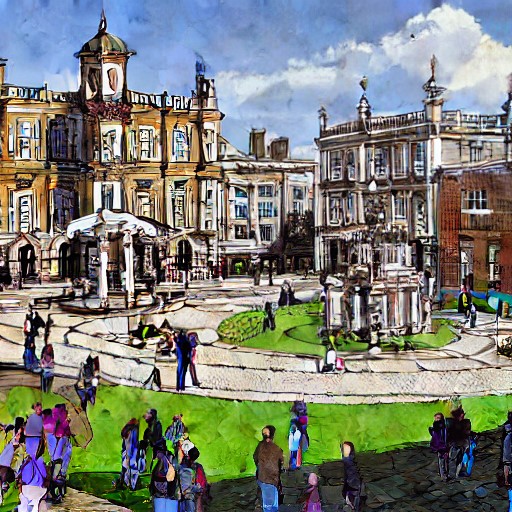} \hspace{-3mm} &
    \includegraphics[width=0.15\linewidth]{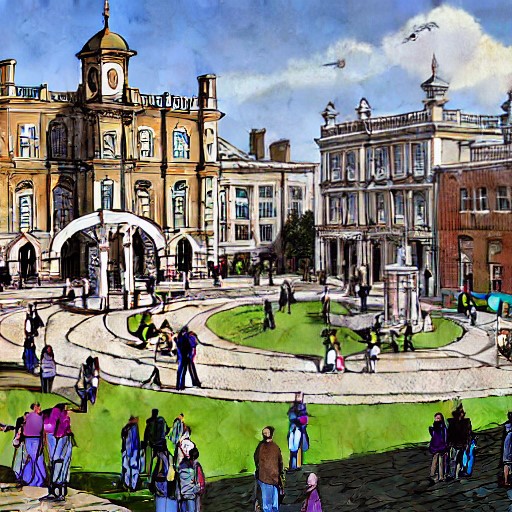} \\

    \multicolumn{2}{c}{(a)} \!\! &
    \multicolumn{2}{c}{(b)} \!\!  &
    \multicolumn{2}{c}{(c)} \\
    
    \includegraphics[width=0.15\linewidth]{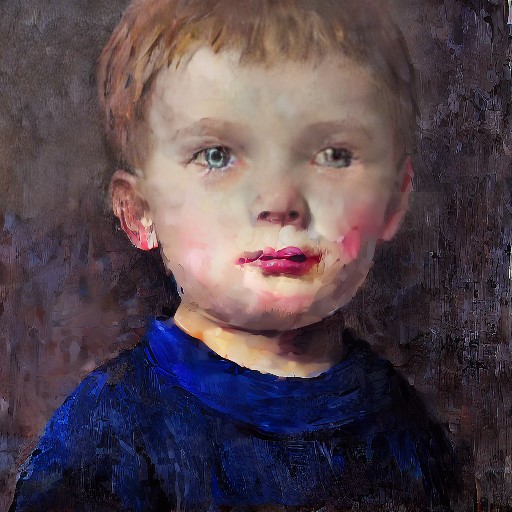} \hspace{-3mm} &
    \includegraphics[width=0.15\linewidth]{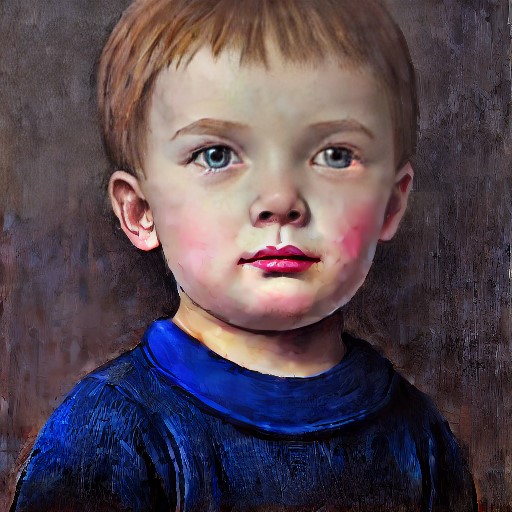}  \!\! &
    \includegraphics[width=0.15\linewidth]{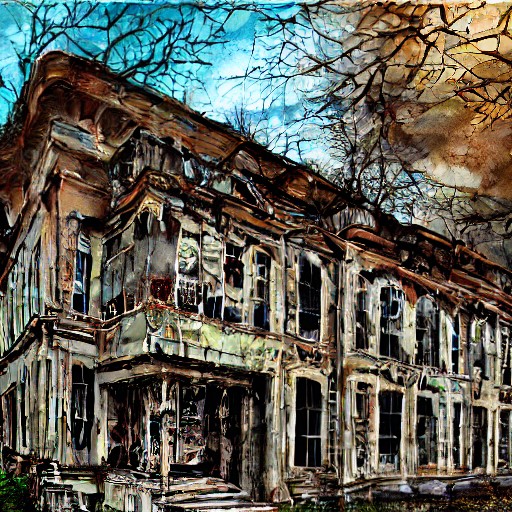}  \hspace{-3mm} &
    \includegraphics[width=0.15\linewidth]{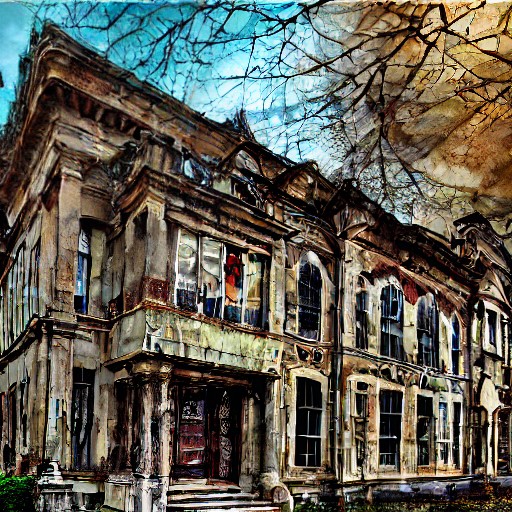}  \!\! &
    \includegraphics[width=0.15\linewidth]{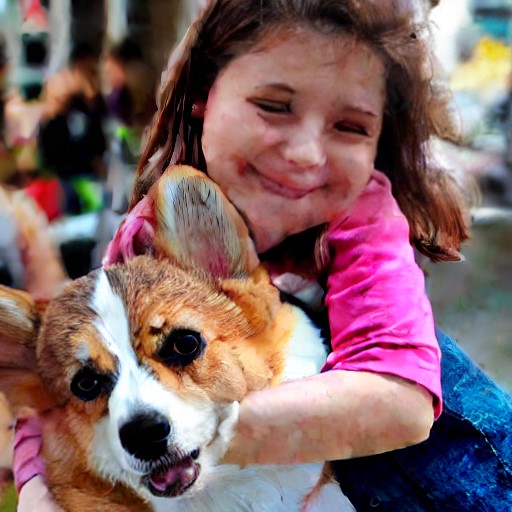}  \hspace{-3mm} &
    \includegraphics[width=0.15\linewidth]{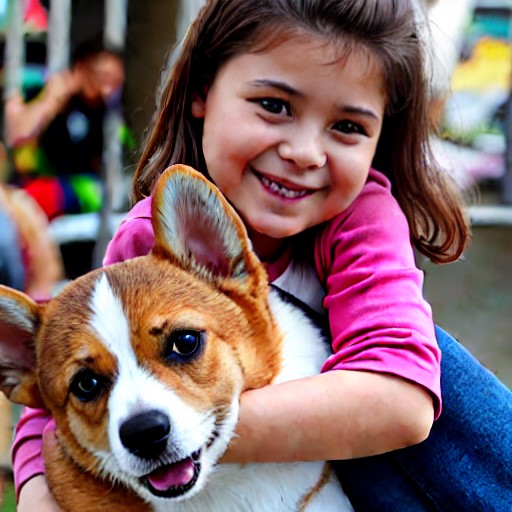} \\

    \multicolumn{2}{c}{(d)} \!\! &
    \multicolumn{2}{c}{(e)} \!\!  &
    \multicolumn{2}{c}{(f)} \\
  \end{tabular}
\vspace{0.2cm}
\captionof{figure}{Segmentation-free vs. classifier-free guidance. Same amount of computations. Best viewed on a computer. Prompts: (a) ``a cute Maltese white dog next to a cat,'' (b) ``ultra realistic, predator, male, fangs, goth, tattoos, leather, fantasy, flesh, bone, body horror, intricate details, eerie, highly detailed, octane render, 8 k, art by artgerm and alphonse mucha and greg rutkowski,'' (c) ``architectural drawing of a new town square for Cambridge England, big traditional museum with columns, fountain in middle, classical design, traditional design, trees,'' (d) ``portrait of a kid,'' (e) ``a beautiful ultradetailed painting of urbex building abandoned, nature, city, unfinished building architecture by april gornik, stormy darkacademia, archdaily, wallpaper, highly detailed, trending on artstation,'' and (f) ``a girl hugging a Corgi on a pedestal.'' }
\vspace{0.2cm}
\label{fig:comp}

\end{center}

}

\begin{document}
\maketitle
\input{sec/0_abstract}    
\input{sec/1_introduction}

\input{sec/2_related_work}
\input{sec/6_background}
\input{sec/3_method}

\input{sec/4_experiments}
\input{sec/5_conclusion}
{
    \small
    \bibliographystyle{ieeenat_fullname}
    \bibliography{main}
}


\end{document}

%% file: preamble.tex
\usepackage[dvipsnames]{xcolor}


%% file: sec/0_abstract.tex
\begin{abstract}

We introduce segmentation-free guidance, a novel method designed for text-to-image diffusion models like Stable Diffusion. Our method does not require retraining of the diffusion model. At no additional compute cost, it uses the diffusion model itself as an \emph{implied} segmentation network, hence named segmentation-free guidance, to dynamically adjust the negative prompt for each patch of the generated image, based on the patch's relevance to concepts in the prompt. We evaluate segmentation-free guidance both objectively, using FID, CLIP, IS, and PickScore, and subjectively, through human evaluators. For the subjective evaluation, we also propose a methodology for subsampling the prompts in a dataset like MS COCO-30K to keep the number of human evaluations manageable while ensuring that the selected subset is both representative in terms of content and fair in terms of model performance. The results demonstrate the superiority of our segmentation-free guidance to the widely used classifier-free method. Human evaluators preferred segmentation-free guidance over classifier-free $60\%$ to $19\%$, with $18\%$ of occasions showing a strong preference. Additionally, PickScore win-rate, a recently proposed metric mimicking human preference, also indicates a preference for our method over classifier-free.


\end{abstract}

%% file: sec/1_introduction.tex
\section{Introduction}
\label{sec:intro}

Diffusion models are powerful generative models for creating visual content from textual prompts. Their success stems from extensive training data and their ability to handle various modalities and signals, enabling diverse applications such as content editing, inpainting, and personalization.

Controlling a diffusion model can be achieved primarily in two ways - \emph{conditioning} and \emph{guidance}. When a diffusion model is conditioned, it is typically trained to accept a particular form of additional conditioning input, such as a text prompt, image edges, segmentation map, and class labels. However, adapting the model to a different condition often necessitates retraining from scratch. This reliance on expensive retraining poses challenges for end-users seeking to adopt and employ conditioning techniques to control diffusion models.


An alternative way to control a diffusion model is through a guidance mechanism. Unlike conditioning techniques, this approach does not rely on an external conditioning signal. Instead, it associates a guidance function with the diffusion model to fulfill a specific target criterion, which could be as simple as minimizing the CLIP distance between the generated image and the provided text description. When sampling an image, the reverse process iterations are steered in the direction of the guidance function's gradient, resulting in constrained image generation. 

When comparing control techniques for diffusion models, guidance emerges as a more versatile approach. It treats the diffusion network as a foundational model which can accommodate different use cases. An earlier method in this domain involved classifier guidance~\cite{dhariwal2021diffusion}, where an explicit classifier functioned as the guidance mechanism. This method utilized the classifier's gradients to drive the image generation process. However, classifier guidance has transitioned to classifier-free guidance~\cite{ho2021classifier}, eliminating the need for an explicit classifier. In classifier-free guidance approaches, the network is trained to adapt class-label information and conditioning signals without relying on a fixed network architecture.

In this paper, we propose enhancing image generation quality beyond classifier-free guidance by introducing a novel and universal segmentation-free guidance approach. This methodology aims to improve image quality of diffusion models without necessitating costly retraining, architectural changes, or additional computing during inference.


Image generation using classifier-free guidance involves two forward passes of the diffusion network per iteration: one that uses conditional information and one that does not. The conditional information generally involves a (positive) text prompt describing different objects of interest in the generated image. For instance, in Fig.~\ref{fig:comp}-a, the positive prompt is "a cute Maltese white dog next to a cat." The forward pass without conditional information is usually carried out by an empty (negative) prompt (i.e., ""). However, it is possible to employ non-empty negative prompts. This type of guidance allows the objects present in the positive, but not the negative, prompt to become more prominent. Nonetheless, the issue with having such negative prompts is that it interacts with the generated image globally. 

Our objective is to dynamically adjust the negative prompt for each image patch. We examine attention maps within the diffusion model, specifically where it interacts with the text prompt embeddings. For each patch of the attention map, we aim to find the object in the positive prompt with the highest correlation. Subsequently, this selected object is excluded from the negative prompt interacting with that specific patch. Accordingly, the forward pass of the diffusion model carries out as if each patch cross-attends dynamically with a different negative prompt.
%
Furthermore, the corresponding attention weight is adjusted to account for self-attention interactions. Since this proposed method of guidance does not involve any segmentation network as a guidance function, we term this method as \emph{segmentation-free guidance}. Our method realizes local interaction between prompt embedding and feature patches while dynamically adjusting the negative prompts; thus, it produces better image generation quality, as shown in Fig.~\ref{fig:comp}. 

Our contributions can be summarized as follows:
\begin{itemize}
    \item We introduce a novel mechanism named segmentation-free guidance that effectively adjusts the negative prompt for each patch of the generated image based on the category of the patch. 
    \item We also propose an efficient subjective evaluation methodology that involves sub-sampling of prompts dataset for assessment. The chosen subset of prompts ensures the representation of dataset diversity while maintaining fairness in terms of model performance. 
    \item Finally, we perform extensive evaluation on the MS-COCO datasets on which we show both qualitative and quantitative improvement. 
\end{itemize}

%% file: sec/2_related_work.tex
\section{Related Work}
\label{sec:related_work}
Our proposed work falls into the scope of controlled image generation using diffusion models. Controlled image generation can be broadly classified into conditional generation and guided generation. These are discussed as follows.
\\\\
\textbf{Conditional Generation}
These category of works generally require training diffusion models from scratch where conditional input can be of the form of prompts~\cite{whang2022deblurring,wang2022semantic,nichol2022glide,ho2021classifier,bansal2022cold}. One of the most popular works~\cite{ho2021classifier} proposed use of classifier-free guidance with class labels as prompts. In this work, the diffusion model is trained such that the output is a linear combination between that of conditional and unconditional outputs. The authors of \cite{bansal2022cold} trained a diffusion model, where it is enforced to solve linear inverse problems. This is realized through a guidance function known as linear degradation operator. \cite{nichol2022glide} used classifier-free guidance but extended it to descriptive phrases as prompts. Furthermore, the network was trained to enforce similarity between CLIP~\cite{radford2021learning} representations of images and text. However, the major disadvantage of conditional generation methods is that the diffusion models need to be retrained and hence it is computationally intensive.
\\\\
\textbf{Guided Generation}
In this category, the diffusion model is kept frozen without any re-training. However, the sampling process for image generation is modified using gradients from a guidance function. There are prior works that studied guided image generation using various constraints and guidance functions~\cite{wang2022zero, lugmayr2022repaint, kawar2022denoising, graikos2022diffusion, dhariwal2021diffusion, chung2022improving, chung2022diffusion}. The most popular method in this category is classifier guidance~\cite{dhariwal2021diffusion}. In this method, a classifier is trained to distinguish images of different scales. The classifier is used as a guidance function, the gradients of which are used in the sampling process. Alternative methods include \cite{wang2022zero}, where the guidance function is a linear operator. Since gradients of the linear operator are used, components of the images were generated in the null space of the linear operator. However, the use of null space does not naturally extend to non-linear guidance functions. In \cite{chung2022diffusion}, the authors did an elaborate analyses of multiple simple non-linear guidance functions, e.g. non-linear blurring. The gradient of the non-linear function was calculated on expected denoised images and the sampling process was modified. Recently, \cite{bansal2023universal} proposed a training-free universal guidance mechanism that can use guidance in the form of CLIP, segmentation map, face recognition, object location, style guidance to produce more controlled image generation.

In this paper, we consider a segmentation-free guidance mechanism, which does not require training from scratch. Furthermore, it does not require extra computation during image sampling compared to classifier-free guidance. Our method modulates the cross-attention weights pertaining to different categories in the prompt, which enhances the visual quality of the generated image.


%% file: sec/6_background.tex
\section{Background}
\label{sec:background}

Gaussian diffusion models \cite{Sohl15, Song19, Ho20} are powerful generative methods for sampling $\mathbf{x}$, e.g., an image, according to $p(\mathbf{x})$, e.g., a dataset. They involve a $T$-step forward diffusion process that creates a Markov chain of ever more noisy latent representations
\begin{equation}
    \mathbf{z}_t = \alpha_t \mathbf{x} + \sigma_t \epsilon, \quad \epsilon \sim \mathcal{N}(\mathbf{0}, \mathbf{I}), \label{eq:0}
\end{equation}
with $\alpha_t^2 = 1/(1 + e^{-\lambda_t})$, $\sigma_t^2 = 1 - \alpha_t^2$,  $\lambda_1 > \dots > \lambda_T$ representing a log-SNR schedule, and a $T$-step reverse denoising process that starts by sampling Gaussian noise $\mathbf{z}_T \sim \mathcal{N}(\mathbf{0}, \mathbf{I})$ and proceeds by sampling $\mathbf{z}_{t-1}$ according to
\begin{equation}
    p_{\theta} (\mathbf{z}_{t-1} |  \mathbf{z}_t) = \mathcal{N}(\mu_{\theta}(\mathbf{z}_t), \Sigma_t ), \nonumber
\end{equation}
where $\mu_{\theta}(\mathbf{z}_t)$ is a function of diffusion model's output $\epsilon_{\theta}(\mathbf{z}_t)$ \cite{ho2021classifier}, which estimates $\epsilon$ in \eqref{eq:0} by training on
\begin{equation}
    \mathbb{E}_{\epsilon, t} [ \lVert \epsilon_{\theta}(\mathbf{z}_t) - \epsilon \rVert_2^2]. \nonumber
\end{equation}
$\epsilon$ and $\epsilon_{\theta}$ are called the true and estimated scores, respectively.

Generative models have successfully been used in many applications, including text-to-image, where it is desired to generate an image consistent with a given prompt $\mathbf{c}$. Generating high quality images, however, requires using guidance methods \cite{dhariwal2021diffusion}. Classifier-free guidance \cite{ho2021classifier} is one such method which is inspired by an \emph{implicit} classifier
\begin{equation}
    \log p^{i}(\mathbf{c}|\mathbf{z}_t) = \log p(\mathbf{z}_t|\mathbf{c}) - \log p(\mathbf{z}_t) + \text{const.}, \label{eq:1}
\end{equation}
with a gradient
\begin{equation}
    \nabla_{\mathbf{z}_t} \log p^{i}(\mathbf{c}|\mathbf{z}_t) \propto - [\epsilon_{\theta}(\mathbf{z}_t, \mathbf{c}) - \epsilon_{\theta}(\mathbf{z}_t) ]. \label{eq:2}
\end{equation}
In classifier-free guidance, during sampling, diffusion model's output $\epsilon_{\theta}(\mathbf{z}_t, \mathbf{c})$ is steered in the direction of \eqref{eq:2} to increase the implicit classifier's log likelihood of \eqref{eq:1} 
\begin{equation}
    \tilde{\epsilon}_{\theta}(\mathbf{z}_t, \mathbf{c}) = (1 + w) \epsilon_{\theta}(\mathbf{z}_t, \mathbf{c}) - w \epsilon_{\theta}(\mathbf{z}_t), \label{eq:3}
\end{equation}
where $w$ is the guidance strength. Note that $\epsilon_{\theta}(\mathbf{z}_t)$ is computed by applying an empty second, a.k.a., negative, prompt, i.e., $\epsilon_{\theta}(\mathbf{z}_t) = \epsilon_{\theta}(\mathbf{z}_t, \varnothing)$. 

%% file: sec/3_method.tex
\section{Segmentation-Free Guidance}
\label{sec:segm_free}

\begin{figure*}
  \centering
  \begin{subfigure}{0.23\linewidth}
    \includegraphics[width=\linewidth]{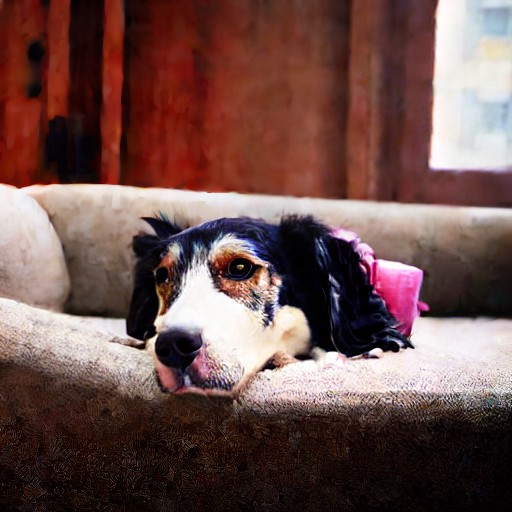}
    \caption{Empty negative prompt, i.e., NP: ""}
    \label{fig:motiv-a}
  \end{subfigure}
  \hfill
  \begin{subfigure}{0.23\linewidth}
    \includegraphics[width=\linewidth]{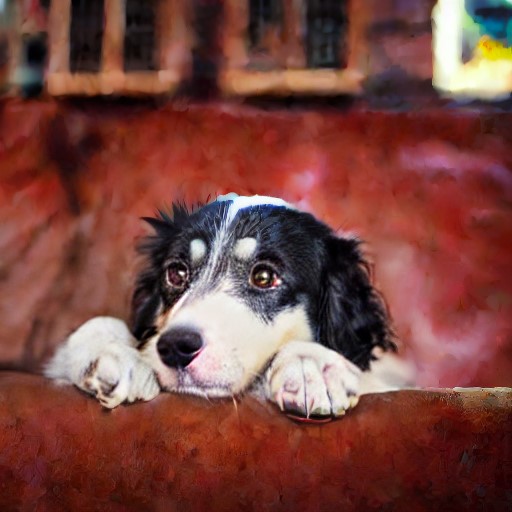}
    \caption{NP: ``A couch in an office" \\ Dog becomes more prominent.}
    \label{fig:motiv-b}
  \end{subfigure}
  \hfill
  \begin{subfigure}{0.23\linewidth}
    \includegraphics[width=\linewidth]{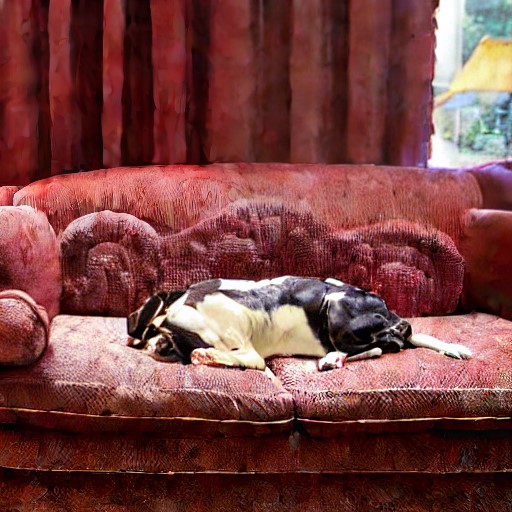}
    \caption{NP: ``A dog in an office" \\ Couch becomes more prominent.}
    \label{fig:motiv-c}
  \end{subfigure}
  \hfill
  \begin{subfigure}{0.23\linewidth}
    \includegraphics[width=\linewidth]{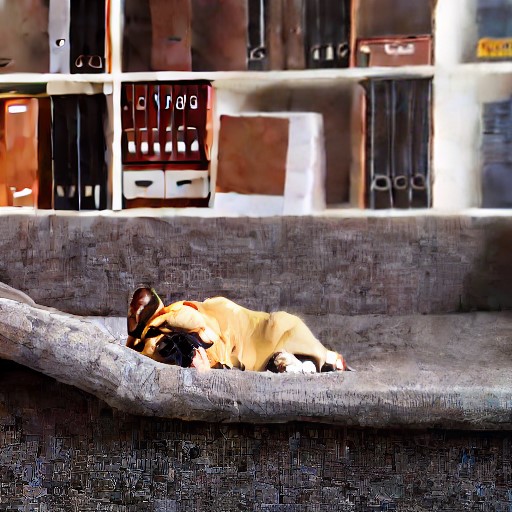}
    \caption{NP: ``A dog on a couch" \\ Office becomes more prominent.}
    \label{fig:motiv-d}
  \end{subfigure}
  \caption{All images are generated using classifier-free guidance from the same seed and positive prompt, ``A dog on a couch in an office," but with different negative prompts (NP). As can be seen, the regions corresponding to the omitted concepts improve in detail.}
  \label{fig:motivation}
\end{figure*}

To illustrate the motivation behind our method, we consider an example. All images in Figure \ref{fig:motivation} were generated using classifier-free guidance from the same seed and positive prompt, "A dog on a couch in an office." However, they differ in the negative prompts used. Figure \ref{fig:motiv-a} employs an empty prompt, while Figures \ref{fig:motiv-b}, \ref{fig:motiv-c}, and \ref{fig:motiv-d} omit the words "dog", "couch", and "office" from the negative prompt, respectively. As evident from the images, the regions corresponding to the omitted concepts exhibit enhanced detail. This raises the question: Can we enhance classifier-free guidance by dynamically adjusting the negative prompt for each patch of the generated image based on its semantic content? Segmentation-free guidance represents one such approach.

Next, we describe our segmentation-free guidance, explaining the motivation behind its different components. Let $\mathbf{c}_i$ denote the $i$-th CLIP embedding of prompt and $\mathbf{z}_p$ the $p$-th patch of $\mathbf{z}_t$. Let us define $\mathbf{z}_p$'s semantic as
\begin{align}
    \mathbf{s}_p \triangleq \argmax_{\mathbf{c}_i} \log p^i(\mathbf{c}_i|\mathbf{z}_p, \mathbf{c} - \{\mathbf{c}_i\}), \label{eq:4}
\end{align}
where $\mathbf{c} - \{\mathbf{c}_i\}$ denotes prompt embeddings with $\mathbf{c}_i$ omitted. The Bayes rule can be used to write (compare to \eqref{eq:1})
\begin{align}
    \log p^i(\mathbf{s}_p|\mathbf{z}_p, \mathbf{c} - \{\mathbf{s}_p\}) = & \log p(\mathbf{z}_p|\mathbf{c}) - \\ \nonumber
    & \log p(\mathbf{z}_p|\mathbf{c}-\{\mathbf{s}_p\}) + \text{const.} \label{eq:5}
\end{align}
Segmentation-free guidance enhances $\mathbf{z}_p$'s detail by steering diffusion model's output $\epsilon_{\theta}(\mathbf{z}_p, \mathbf{c}$) during sampling, in the direction that increases $\mathbf{s}_p$'s log-likelihood
\begin{eqnarray}
    \nabla_{\mathbf{z}_p} \log p^i(\mathbf{s}_p|\mathbf{z}_p, \mathbf{c} - \{ \mathbf{s}_p \}) \propto & \nonumber \\ & \hspace{-20mm} - [\epsilon_{\theta}(\mathbf{z}_p, \mathbf{c}) - \epsilon_{\theta}(\mathbf{z}_p, \mathbf{c} - \{ \mathbf{s}_p \}) ]. \label{eq:6}
\end{eqnarray}
In \eqref{eq:6}, diffusion model's output $\epsilon_{\theta}(\mathbf{z}_p, \mathbf{c})$ has been used to estimate the true score $\epsilon^*(\mathbf{z}_p, \mathbf{c})$. Since $\mathbf{s}_p$ is not known a priori, $\epsilon_{\theta}(\mathbf{z}_p, \mathbf{c} - \{\mathbf{c}_i\})$ has to be computed for all $i$, which makes \eqref{eq:6} computationally prohibitive.

To overcome this limitation, we modify \eqref{eq:6} to use $\mathbf{z}_p$'s local (i.e., layer) semantic
\begin{eqnarray}
    \nabla_{\mathbf{z}_p} \log p^i(\mathbf{s}_p|\mathbf{z}_p, \mathbf{c} - \{ \mathbf{s}_p \}) \propto & \nonumber \\ & \hspace{-20mm} - [\epsilon_{\theta}(\mathbf{z}_p, \mathbf{c}) - \epsilon_{\theta}(\mathbf{z}_p, \mathbf{c} - \{ \mathbf{s}_{pl} \}_{l=1}^L) ]. \label{eq:7}
\end{eqnarray}
defined as (compare to \eqref{eq:4})
\begin{equation}
    \mathbf{s}_{pl} \triangleq \argmax_{\mathbf{c}_i, i > 0} A(\mathbf{z}_{pl}, \mathbf{c}_i), \label{eq:8}
\end{equation}
where $l$ indexes diffusion model's cross-attention modules and $\mathbf{z}_{pl}$ denotes their input patch. $A(\mathbf{z}_{pl}, \mathbf{c}_i)$ denotes the computed cross-attention weight. As shown in Section \ref{sec:experiments}, segmentation-free guidance offers a gain justifying use of $\mathbf{s}_{pl}$ instead of $\mathbf{s}_{p}$; However, as a motivation, we note that diffusion models like Stable Diffusion have an architecture similar to segmentation networks like Mask2Former \cite{cheng2022masked}, i.e., they are composed of a stack of transformer decoder layers. This suggests that $\mathbf{s}_{pl}$ could serve as a proxy for $\mathbf{s}_p$, though there is no guarantee of consistent semantics for a patch across layers, or for adjacent patches from the same layer. This is especially true during the first few iterations because of noise. Hence, we use classifier-free guidance for the first few iterations and switch to segmentation-free guidance only after, e.g., $t > t_s=10$. We also note that in \eqref{eq:8}, $\mathbf{c}_0$, i.e., begin-of-sentence (BOS) embedding, has been excluded from designation as a patch's local semantic. This is because BOS's cross-attention weight is almost always significantly larger than those of others prompt tokens, while its projected value is negligible, i.e., BOS serves as a neutral text embedding.

Finally, we note that conditioning leakage, i.e., $\mathbf{s}_{pl}$ indirectly affecting $\epsilon_{\theta}(\mathbf{z}_p, \mathbf{c} - \{ \mathbf{s}_{pl} \}_{l=1}^L)$, negatively affects guidance according to \eqref{eq:7}: First, we note that $\mathbf{c} - \{\mathbf{c}_i\}$ is different from the CLIP text encoding of a prompt that omits the $i$-th token. In the former the $i$-th token still substantially affects the subsequent embeddings $\{ \mathbf{c}_j \}_{j>i}$. The second source of leakage are model's self-attention modules that allow a patch to be indirectly affected by conditioning applied to other patches. As a result of conditioning leakage, the gradients given by \eqref{eq:7} are small and ineffective. Therefore we further modify it to
\begin{eqnarray}
    \nabla_{\mathbf{z}_p} \log p^i(\mathbf{s}_p|\mathbf{z}_p, \mathbf{c} - \{\mathbf{s}_p\}) \propto & \nonumber \\ & \hspace{-20mm} - [\epsilon_{\theta}(\mathbf{z}_p, \mathbf{c}) - \bar{\epsilon}_{\theta}(\mathbf{z}_p, \mathbf{c}) ], \label{eq:9}
\end{eqnarray}
where $\bar{\epsilon}_{\theta}(\mathbf{z}_{\lambda}, \mathbf{c})$ is computed identically to $\epsilon_{\theta}(\mathbf{z}_p, \mathbf{c})$, with the difference that in each cross-attention module, the $\mathbf{s}_{pl}$'s attention weight (c.f. \eqref{eq:8}) is multiplied by $-a$, where $a$ is a positive number, e.g., $10$, called the segmentation-free scale. This effectively compensates for the conditioning leakage. Thus the segmentation-free modified score is 
\begin{equation}
    \tilde{\epsilon}(\mathbf{z}_t, \mathbf{c}) = (1 + \bar{w})\epsilon(\mathbf{z}_t, \mathbf{c}) - \bar{w} \bar{\epsilon}(\mathbf{z}_t),
\end{equation}
where $\bar{w}$ denotes segmentation-free guidance scale. We have found, through experiments, that a smaller value of $\bar{w} = 2.5$ (compared to $w=7.5$) gives very good results. Note that segmentation-free guidance has almost the same computational complexity as classifier-free. \cref{alg:segm-free} gives the segmentation-free guidance method.

\setlength{\textfloatsep}{10pt} 
\begin{algorithm}[h]
\small
\SetAlgoLined

 \textbf{Require:} $w$: Classifier-free guidance strength ($7.5$) \\
 \textbf{Require:} $T$: Total number of iterations ($20$) \\
 \textbf{Require:} $t_s$: Classifier-free guidance iterations ($T/2$) \\
 \textbf{Require:} $a$: Segmentation-free scale (10.0) \\
 \textbf{Require:} $\bar{w}$: Segmentation-free guidance strength (2.5) \\
 \textbf{Require:} $\mathbf{c}$: Prompt CLIP text embeddings \\
 \textbf{Require:} $\lambda_1 > \cdots \lambda_T$: log-SNR schedule \\
 
 \quad 01: $\mathbf{z}_T \sim \mathcal{N}(\mathbf{0}, \mathbf{I})$ \\
 \quad 02: \textbf{for}  $t = T, \cdots, 1$  \textbf{do} \\
 \quad 03: \quad \textbf{if} $t \geq T-t_s$ \\
 \quad \quad \quad \quad \quad \# compute classifier-free score \\
 \quad 04: \quad \quad $\tilde{\epsilon}(\mathbf{z}_t, \mathbf{c}) = (1 + w)\epsilon(\mathbf{z}_t, \mathbf{c}) - w \epsilon(\mathbf{z}_t)$ \\
 \quad 05: \quad \textbf{else} \\
 \quad \quad \quad \quad \quad \# compute segmentation-free score \\
 \quad 06: \quad \quad $\tilde{\epsilon}(\mathbf{z}_t, \mathbf{c}) = (1 + \bar{w})\epsilon(\mathbf{z}_t, \mathbf{c}) - \bar{w} \bar{\epsilon}(\mathbf{z}_t)$ \\
 \quad \quad \quad \# sample $\mathbf{z}_{t-1}$ \\
 \quad 07: \quad \textbf{if} $t > 1$ \\
 \quad 08: \quad \quad $\mathbf{z}_{t-1} \sim \mathcal{N}(\mu_{\theta}(\mathbf{z}_t), \Sigma_t )$ \\
 \quad 09: \quad \textbf{else} \\
 \quad 10: \quad \quad  $\mathbf{z}_0 = (\mathbf{z}_1 - \sigma_1 \tilde{\epsilon_1}) / \alpha_1$ \\
 \quad 11: \textbf{return} $\mathbf{x} = \mathbf{z}_0$ \\
\caption{Segmentation-free guidance}
\label{alg:segm-free}
\end{algorithm}

%% file: sec/4_experiments.tex
\section{Experiments}
\label{sec:experiments}
In this section we report our experiments. We provide detailed analysis of a few cases in \cref{sec:case_studies}, give our quantitative results in \cref{sec:objective} and finally report our qualitative results in \cref{sec:subjective}. All images and results are generated using the Stable Diffusion v$1.5$ model with $T = 20$. Unless otherwise stated, classifier-free guidance strength is set to $w = 7.5$. We use a segmentation-free guidance strength of $\Bar{w} = 2.5$ (c.f., \cref{alg:segm-free}).

\subsection{Case Studies} \label{sec:case_studies}
To provide insight into the nature of the improvement and the role of different parameters, we present a few case studies comparing images generated using the two guidance methods. We also include some examples showing segmentation-free method's limitations.

\begin{figure}
  \centering
  \begin{subfigure}{0.48\linewidth}
    \includegraphics[width=\linewidth]{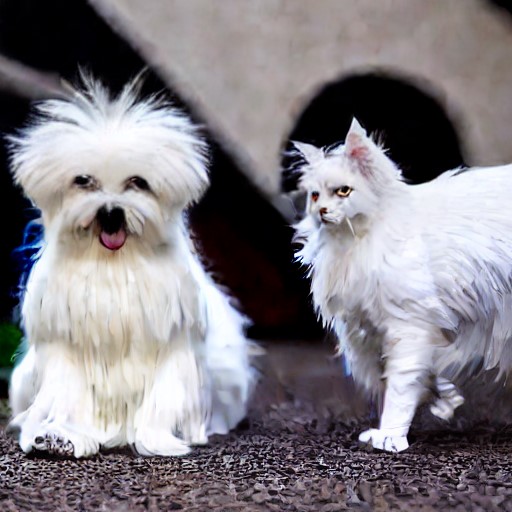}
    \caption{Classifier-free guidance}
    \label{fig:effect_segm_free-a}
  \end{subfigure}
  \hfill
  \begin{subfigure}{0.48\linewidth}
    \includegraphics[width=\linewidth]{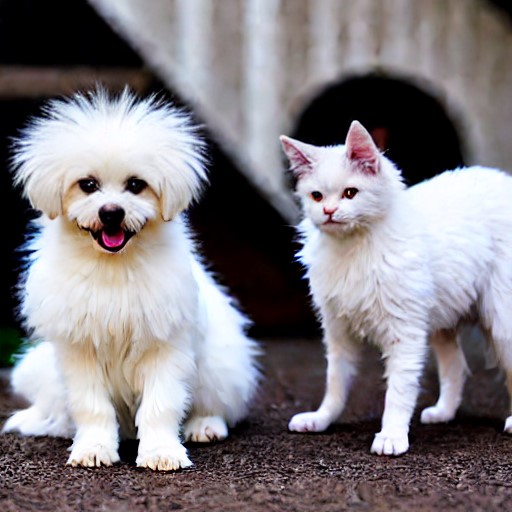}
    \caption{Segmentation-free, $t_s = 5$}
    \label{fig:effect_segm_free-b}
  \end{subfigure}
  \hfill
  \caption{Effect of segmentation-free guidance. Prompt: ``a cute Maltese white dog next to a cat". The long hair characteristic of the dog spills over to the cat under classifier-free guidance (\ref{fig:effect_segm_free-a}). Segmentation-free guidance improves quality by adjusting conditioning for the dog and the cat individually (\ref{fig:effect_segm_free-b}).}
  \label{fig:effect_segm_free}
\end{figure}

\cref{fig:effect_segm_free} shows images generated using classifier-free and segmentation-free guidance methods for the prompt "a cute Maltese white dog next to a cat". As \cref{fig:effect_segm_free-a} shows, under classifier-free guidance some of the dog's features, particularly its long white hair, interfere with cat's rendering, resulting in its unnatural appearance. Segmentation-free guidance improves details of the dog and the cat by adjusting their conditioning individually (\ref{fig:effect_segm_free-b}).

\begin{figure*}
  \centering
  \begin{subfigure}{0.19\linewidth}
    \includegraphics[width=\linewidth]{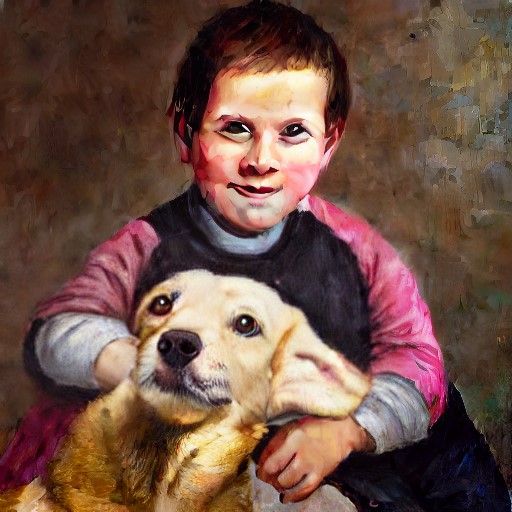}
    \caption{Classifier-free}
    \label{fig:effect_a-a}
  \end{subfigure}
  \hfill
  \begin{subfigure}{0.19\linewidth}
    \includegraphics[width=\linewidth]{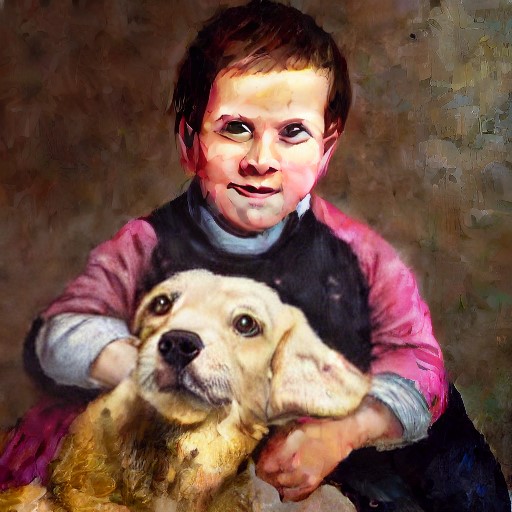}
    \caption{Segm.-free, $a = 0$}
    \label{fig:effect_a-b}
  \end{subfigure}
  \hfill
  \begin{subfigure}{0.19\linewidth}
    \includegraphics[width=\linewidth]{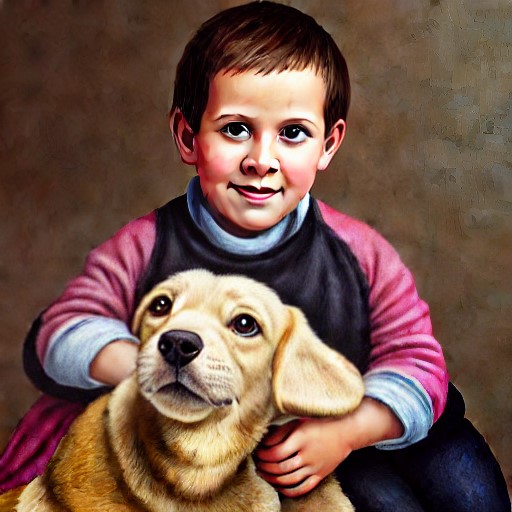}
    \caption{Segm.-free, $a = 10$}
    \label{fig:effect_a-c}
  \end{subfigure}
  \hfill
  \begin{subfigure}{0.19\linewidth}
    \includegraphics[width=\linewidth]{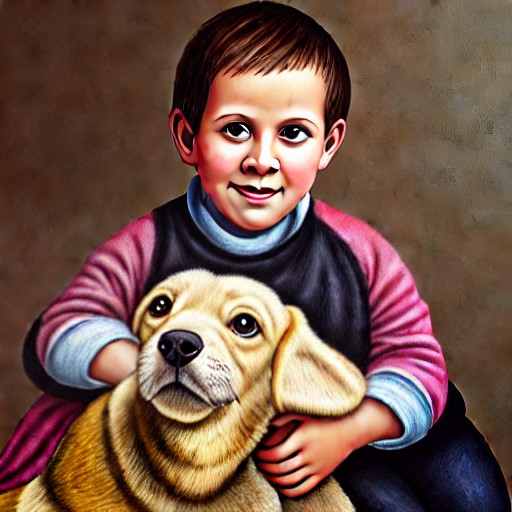}
    \caption{Segm.-free, $a = 20$}
    \label{fig:effect_a-d}
  \end{subfigure}
  \hfill
  \begin{subfigure}{0.19\linewidth}
    \includegraphics[width=\linewidth]{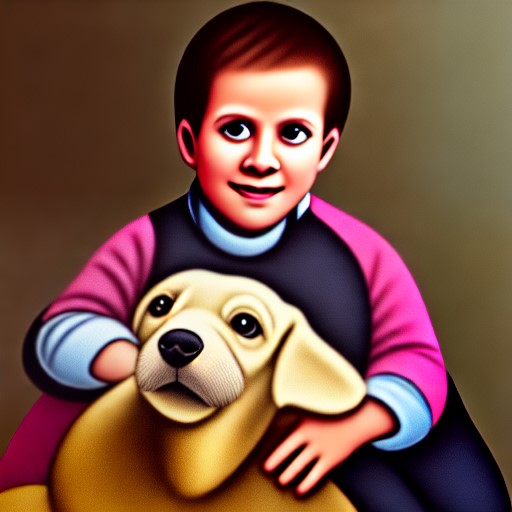}
    \caption{Segm.-free, $a = 50$}
    \label{fig:effect_a-e}
  \end{subfigure}
  \caption{Effect of segmentation-free scale ($a$ in \cref{alg:segm-free}). Prompt: "portrait of a dog and a kid". $a = 0$ is ineffective due to conditioning leakage (\ref{fig:effect_a-b}), while large values $a > 20$ cause artifacts (\ref{fig:effect_a-d}, \ref{fig:effect_a-e}). Subjective evaluations show that $a = 10$ gives the best results (\ref{fig:effect_a-c}).}
  \label{fig:effect_a}
\end{figure*}

\cref{fig:effect_a} shows the effect of segmentation-free scale (parameter $a$ in \cref{alg:segm-free}). As evident from \cref{fig:effect_a-b}, setting $a = 0$, which is equivalent to ignoring the most relevant prompt CLIP embedding, is ineffective. As explained in \cref{sec:segm_free}, this is due to conditioning leakage resulting from both, correlation among prompt text embeddings, and self-attention modules in the diffusion model. In contrast, as seen in \cref{fig:effect_a-d} and \ref{fig:effect_a-e}, very large values, e.g., $a > 20$, cause artifacts by over compensating for this leakage. In \cref{sec:subjective} we present human evaluation results showing $a = 10$ to provide the best results.

\begin{figure}
  \centering
  \begin{subfigure}{0.32\linewidth}
    \includegraphics[width=\linewidth]{fig/corgi_seed1_cls7p5.jpg}
    \caption{Classifier-free guidance}
    \label{fig:effect_ts_comp-a}
  \end{subfigure}
  \hfill
  \begin{subfigure}{0.32\linewidth}
    \includegraphics[width=\linewidth]{fig/corgi_seed1_segms10t10.jpg}
    \caption{Segmentation-free, $t_s = 10$}
    \label{fig:effect_ts_comp-b}
  \end{subfigure}
  \hfill
  \begin{subfigure}{0.32\linewidth}
    \includegraphics[width=\linewidth]{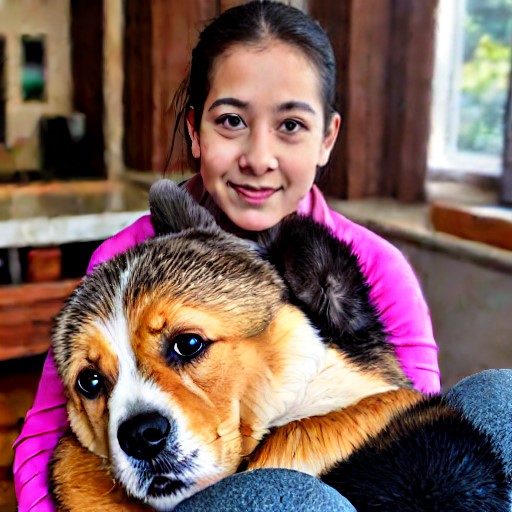}
    \caption{Segmentation-free, $t_s = 1$}
    \label{fig:effect_ts_comp-c}
  \end{subfigure}
  \caption{Compositional effect of $t_s$. Prompt: "a girl hugging a Corgi on a pedestal". Switching too early from classifier-free to segmentation-free guidance improves local detail, but hurts the overall composition, i.e., too large a dog in (\ref{fig:effect_ts_comp-c}). }
  \label{fig:effect_ts_comp}
\end{figure}

\begin{figure}
  \centering
  \begin{subfigure}{0.32\linewidth}
    \includegraphics[width=\linewidth]{fig/town_seed1_cls7p5.jpg}
    \caption{Classifier-free guidance}
    \label{fig:effect_ts_ignor-a}
  \end{subfigure}
  \hfill
  \begin{subfigure}{0.32\linewidth}
    \includegraphics[width=\linewidth]{fig/town_seed1_segms10t10.jpg}
    \caption{Segmentation-free, $t_s = 10$}
    \label{fig:effect_ts_ignor-b}
  \end{subfigure}
  \hfill
  \begin{subfigure}{0.32\linewidth}
    \includegraphics[width=\linewidth]{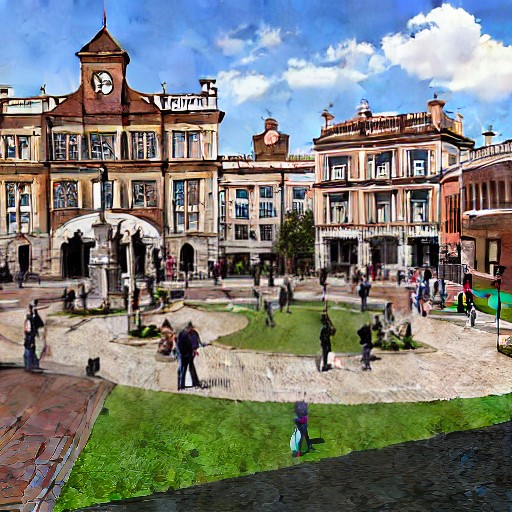}
    \caption{Segmentation-free, $t_s = 5$}
    \label{fig:effect_ts_ignor-c}
  \end{subfigure}
  \caption{Effect of $t_s$ in skipping aspects of prompt. Prompt: "architectural drawing of a new town square for Cambridge England, big traditional museum with columns, fountain in middle, classical design, traditional design, trees". Using a smaller $t_s$ in (\ref{fig:effect_ts_ignor-c}) improves the overall image quality with respect to (\ref{fig:effect_ts_ignor-b}), but at the expense of ignoring the "architectural drawing" aspect of the prompt.}
  \label{fig:effect_ts_ignor}
\end{figure}

We next consider the effects of the number of classifier-free guidance iterations performed before switching to segmentation-free guidance (parameter $t_s$ in \cref{alg:segm-free}). Generally, reducing $t_s$ improves a region's detail by customizing its conditioning earlier. However, switching too early may have some negative effects. For example, \cref{fig:effect_ts_comp-c} shows images generated for the prompt "a girl hugging a Corgi on a pedestal". As \cref{fig:effect_ts_comp-b} shows, segmentation-free guidance with $t_s = 10$ greatly enhances the image. However, reducing $t_s$ to $1$ in \cref{fig:effect_ts_comp-c} causes compositional defects, i.e., the dog becomes too large. \cref{fig:effect_ts_ignor} shows another negative effect of using a small $t_s$. While segmentation-free guidance with $t_s = 10$ enhances the image in \cref{fig:effect_ts_ignor-b}, using $t_s = 5$ in \cref{fig:effect_ts_ignor-c} has caused prompt's "architectural drawing" directive to be ignored. We note, however, that using larger values of classifier-free guidance strength $w$ generally requires using a smaller $t_s$, as argued next.

\begin{figure}
  \centering
  \begin{subfigure}{0.32\linewidth}
    \includegraphics[width=\linewidth]{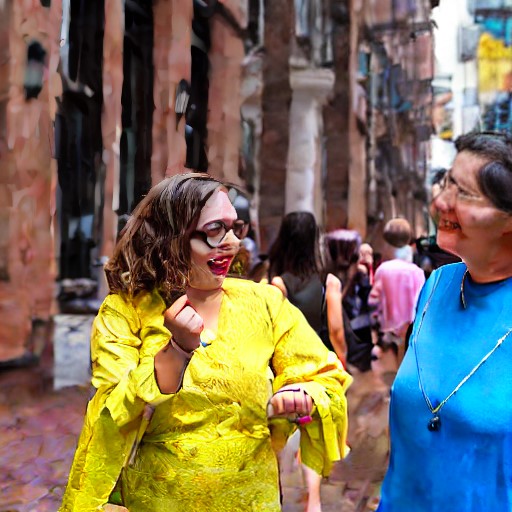}
    \caption{Seed $1$}
    \label{fig:effect_clip-a}
  \end{subfigure}
  \hfill
  \begin{subfigure}{0.32\linewidth}
    \includegraphics[width=\linewidth]{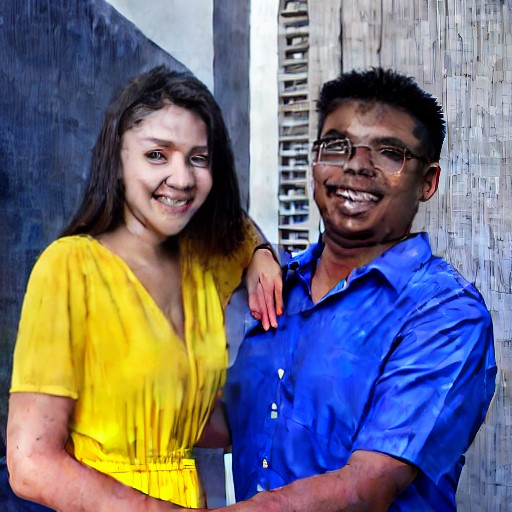}
    \caption{Seed $2$}
    \label{fig:effect_clip-b}
  \end{subfigure}
  \hfill
  \begin{subfigure}{0.32\linewidth}
    \includegraphics[width=\linewidth]{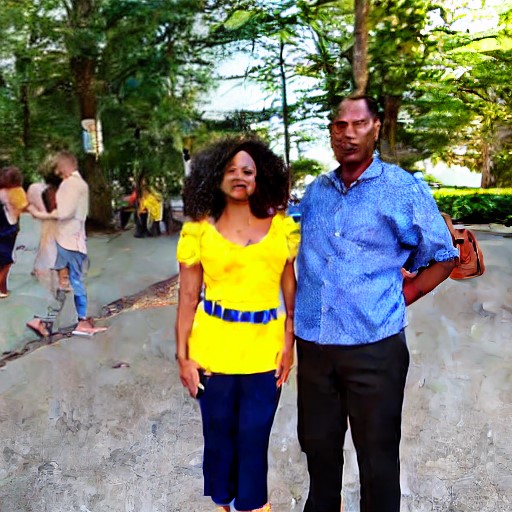}
    \caption{Seed $3$}
    \label{fig:effect_clip-c}
  \end{subfigure}
  \caption{Effect of CLIP text encoding. All images are for the prompt "A man in yellow shirt next to a woman in blue dress" and generated from three different seeds. Note that in (\ref{fig:effect_clip-b}, \ref{fig:effect_clip-c}) the colors are switched, while in (\ref{fig:effect_clip-a}) there is no man. This may be due to "yellow shirt" coming before "woman" in the prompt, whereas "blue dress" comes after it (CLIP is a unidirectional text encoder), or the fact that men with blue shirts appear more often in Stable Diffusion's training data than yellow ones.}
  \label{fig:effect_clip}
\end{figure}

\begin{figure*}
  \centering
  \begin{subfigure}{0.23\linewidth}
    \includegraphics[width=\linewidth]{fig/hybrid_seed1_couple_cls7p5.jpg}
    \caption{Classifier-free, $w = 7.5$}
    \label{fig:effect_w-a}
  \end{subfigure}
  \hfill
  \begin{subfigure}{0.23\linewidth}
    \includegraphics[width=\linewidth]{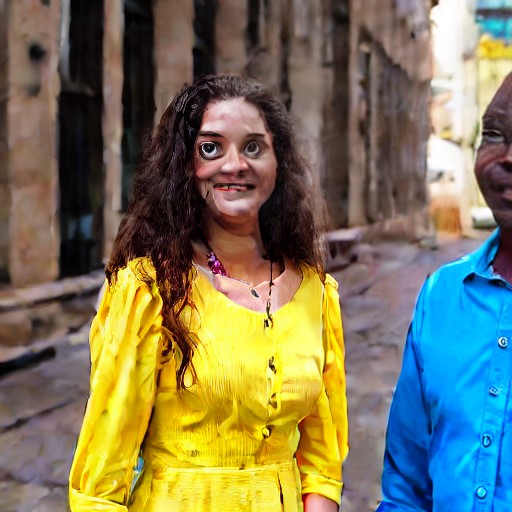}
    \caption{Classifier-free, $w = 12.5$}
    \label{fig:effect_w-b}
  \end{subfigure}
  \hfill
  \begin{subfigure}{0.23\linewidth}
    \includegraphics[width=\linewidth]{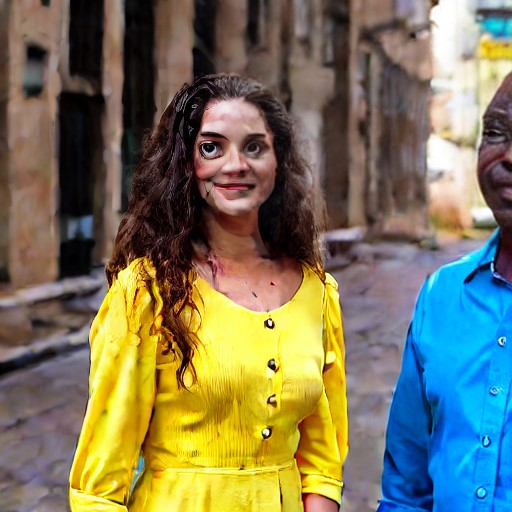}
    \caption{Segm.-free, $t_s = 10$}
    \label{fig:effect_w-c}
  \end{subfigure}
  \hfill
  \begin{subfigure}{0.23\linewidth}
    \includegraphics[width=\linewidth]{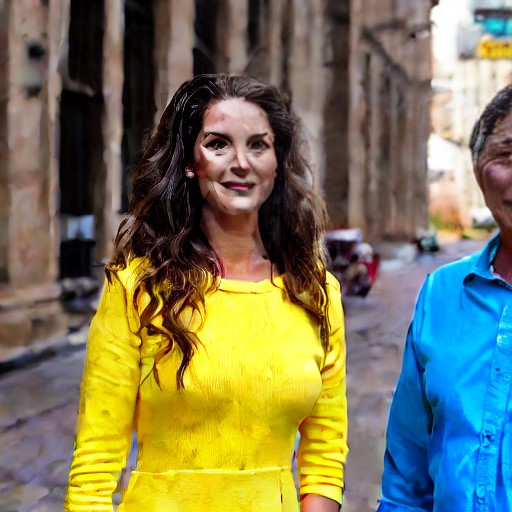}
    \caption{Segm.-free, $t_s = 5$}
    \label{fig:effect_w-d}
  \end{subfigure}
  \hfill
  \caption{A larger $w$ requires a smaller $t_s$. All images generated for the prompt "A man in yellow shirt next to a woman in blue dress". A larger $w = 12.5$ aligns \ref{fig:effect_w-b} better with the prompt (a man appears, although colors are still switched). However the larger $w$ means more aggressive suppression of unrelated noise, which makes a moderate $t_s = 10$ ineffective (\ref{fig:effect_w-c}). A smaller $t_s = 5$ fixes this issue in (\ref{fig:effect_w-d}).}
  \label{fig:effect_w}
\end{figure*}

\begin{figure}
  \centering
  \begin{subfigure}{0.48\linewidth}
    \includegraphics[width=\linewidth]{fig/hybrid_seed1_couple_cls12p5_segms10t5.jpg}
    \caption{$t = 20 (= T)$}
    \label{fig:effect_noise-a}
  \end{subfigure}
  \hfill
  \begin{subfigure}{0.48\linewidth}
    \includegraphics[width=\linewidth]{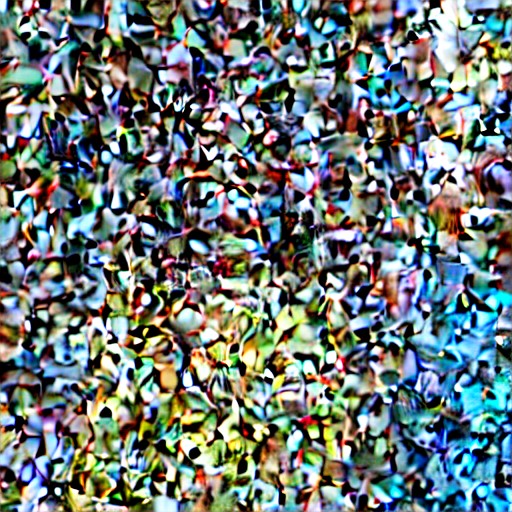}
    \caption{$t = 5 (= t_s)$}
    \label{fig:effect_noise-b}
  \end{subfigure}
  \hfill
  \caption{Importance of using the diffusion model for implicit segmentation. (\ref{fig:effect_noise-b}) is the intermediate image after $t_s = 5$ iterations of classifier-free guidance, and before applying segmentation-free guidance which results in (\ref{fig:effect_noise-a}) at the end. As can be seen, the image is too noisy to be processed by any off-the-shelf network.}
  \label{fig:effect_noise}
\end{figure}

It is a well known fact that the performance of text-to-image diffusion models, e.g., Stable Diffusion, is very dependent on the quality of the text encoder used, e.g., CLIP \cite{saharia2022photorealistic}. For example \cref{fig:effect_clip} shows images generated from three different seeds for the prompt "A man in yellow shirt next to a woman in blue dress". Note that in \cref{fig:effect_clip-b} and \ref{fig:effect_clip-c} the colors are switched, while in \cref{fig:effect_clip-a}, there is no man. The first issue may either be due to "yellow shirt" coming before "woman" in the prompt, compared to "blue dress" that is coming after (CLIP is a unidirectional text encoder), or the fact that men wearing blue shirts appear more often in Stable Diffusion's training data than yellow ones. In either case segmentation-free guidance cannot rectify this issue as it relies on classifier-free guidance for the first few iterations. However, the second issue, i.e., \cref{fig:effect_clip-a} not showing a man, can be rectified by using a larger classifier-free guidance strength, i.e., $w=12.5$ as shown in \cref{fig:effect_w-b}. A larger $w$, however, suppresses unrelated noise more aggressively, which makes a moderate value of $t_s = 10$ ineffective as seen in \cref{fig:effect_w-c}. A smaller $t_s = 5$ however provides a significant improvement, c.f., \cref{fig:effect_w-d}. We show the noisy image resulting after applying $t_s = 5$ iterations of classifier-free guidance in \cref{fig:effect_noise-b}. This is the image that is used in the first iteration of segmentation-free guidance. As expected, the image is quite noisy, which shows the importance of relying on the diffusion model as an implicit segmentation network.

\begin{figure}
  \centering
  \begin{subfigure}{0.32\linewidth}
    \includegraphics[width=\linewidth]{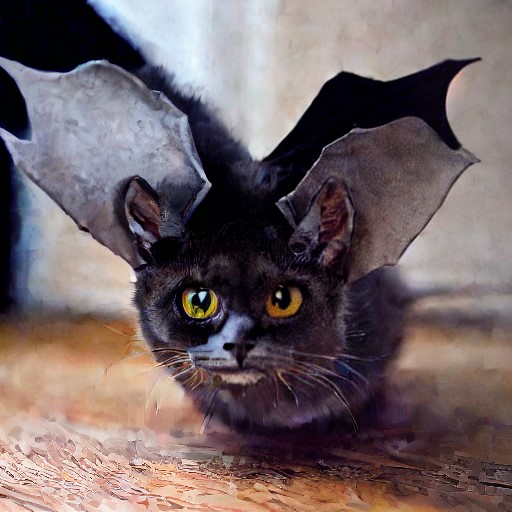}
    \caption{Classifier-free guidance}
    \label{fig:effect_hybrid-a}
  \end{subfigure}
  \hfill
  \begin{subfigure}{0.32\linewidth}
    \includegraphics[width=\linewidth]{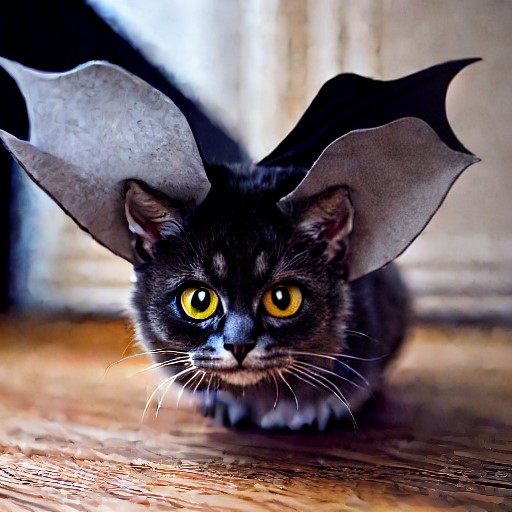}
    \caption{Segmentation-free, $t_s = 10$}
    \label{fig:effect_hybrid-b}
  \end{subfigure}
  \hfill
  \begin{subfigure}{0.32\linewidth}
    \includegraphics[width=\linewidth]{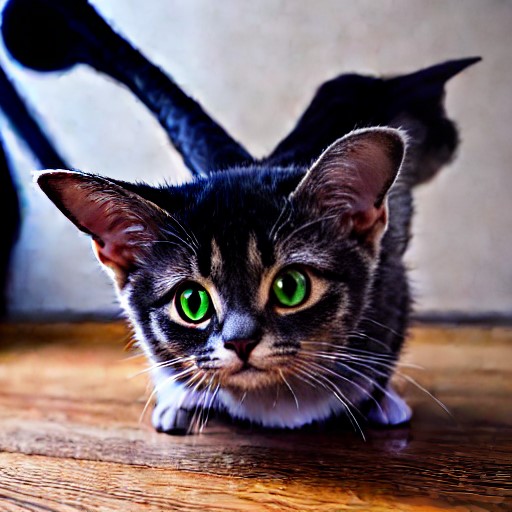}
    \caption{Segmentation-free, $t_s = 5$}
    \label{fig:effect_hybrid-c}
  \end{subfigure}
  \hfill
  \caption{A limitation of segmentation-free guidance. Prompt: "hybrid of a bat and a cat". Segmentation-free guidance (\ref{fig:effect_hybrid-b}) under-performs classifier-free (\ref{fig:effect_hybrid-a}) when the goal is to morph distinct concepts into one. Reducing $t_s$ (\ref{fig:effect_hybrid-c}) exacerbates the deficit, as expected. }
  \label{fig:effect_hybrid}
\end{figure}

Finally we note that segmentation-free guidance improves on classifier-free method by identifying the most relevant concept from the prompt for each patch and excluding the rest from interfering. This bring us to the interesting case where morphing distinct concepts in prompt, such as the "bat" and the "cat" in the prompt "hybrid of a bat and a cat" is actually the goal. As \cref{fig:effect_hybrid} shows, segmentation-free under-performs classifier-free guidance, and reducing $t_s$ exacerbates the deficiency.

\subsection{Quantitative Results} \label{sec:objective}

\textbf{Evaluation metrics}. We report our results on the MS-COCO dataset~\cite{lin2014microsoft}. Following prior works ~\cite{ramesh2022hierarchical,saharia2022photorealistic,balaji2022ediffi}, we utilize the first $30$K captions from the val2014 subset for image generation (also known as MS-COCO-$30$K). A variety of evaluation metrics are adopted to measure the objective quality of these synthesized images, including FID~\cite{heusel2017gans} score, CLIP score~\cite{hessel2021clipscore}, IS score~\cite{salimans2016improved}, and PickScore~\cite{Kirstain2023PickaPicAO} . We compute the FID score between the $30$K generated images and $30$K reference ground truth images following the official implementation, which leverages the \texttt{Inception-V3} model. In line with recent studies~\cite{ramesh2022hierarchical,saharia2022photorealistic,balaji2022ediffi}, we use \texttt{ViT-g-14} model for computing the CLIP score. We calculate the IS score using \texttt{Inception-V3} model. For PickScore, we follow the official implementation \cite{kirstain2023pickapic} (which uses \texttt{CLIP-ViT-H-14} model) and report the win-rate. As stated earlier, all images are generated using the Stable Diffusion v$1.5$ model with $T = 20$. Unless otherwise stated, classifier-free guidance strength is set to $w = 7.5$ and a segmentation-free guidance strength of $\Bar{w} = 2.5$.

\cref{tab:fid_scores} gives the FID, CLIP and IS scores for classifier-free and segmentation-free guidance methods. As the table shows, segmentation-free guidance does not improve either FID or CLIP scores. There is increasing evidence \cite{kirstain2023pickapic, podell2023sdxl, betzalel2022}, however, that such metrics may not reflect visual aesthetics. In fact \cite{kirstain2023pickapic}, based on their large dataset of text-to-image prompts and human preferences, finds that FID is negatively correlated with subjective quality. \cite{kirstain2023pickapic} further leverages this dataset to train a scoring function, PickScore, which predicts human preferences well. \cref{tab:pick_scores} gives the PickScore win-rate for segmentation-free guidance against classifier-free. As the table shows, segmentation-free guidance, with scales $a=5$ and $10$, score PickScore win rates of $63.24\%$ and $60.25\%$ against classifier-free, respectively.

\begin{table}
  \centering
  \begin{tabular}{l c c c}
    \toprule
    Method & FID$\downarrow$ & CLIP$\uparrow$ & IS$\uparrow$ \\
    \midrule
    Class.-free & $17.37$ & $0.3044$ & $37.51$ \\
    Segm.-free, $a=5$ & $19.47$ & $0.2982$ & $37.96$ \\
    Segm.-free, $a=10$ & $20.55$ & $0.2961$ & $25.59$ \\
    \bottomrule
  \end{tabular}
  \caption{FID, CLIP and IS scores for classifier-free and segmentation-free guidance methods.}
  \label{tab:fid_scores}
\end{table}

\begin{table}
  \centering
  \begin{tabular}{l c}
    \toprule
    Method vs. Class.-free & PickScore win-rate \\
    \midrule
    Segm.-free, $a=5$ & $63.24\%$ \\
    Segm.-free, $a=10$ & $60.25\%$ \\
    \bottomrule
  \end{tabular}
  \caption{PickScore win-rate for segmentation-free guidance against classifier-free.}
  \label{tab:pick_scores}
\end{table}

\subsection{Qualitative Results} \label{sec:subjective}
To further demonstrate the performance of our segmentation-free guidance, we conduct a subjective evaluation study. We use the MS-COCO-$30$K validation set prompts to generate images using the two guidance methods and ask human evaluators to choose one of five options based on image quality and match to the prompts. The five options are "much better," "slightly better," "no preference," "slightly worse," and "much worse." Evaluating on the entire MS-COCO-$30$K dataset requires a large number of human evaluations, so we sample the dataset to form a smaller subset. This raises a few important considerations: the size of the subset, how to ensure that it adequately captures the dataset's diversity, and how to ensure that it represents the diffusion model's performance. 

\begin{figure}
  \centering
    \includegraphics[width=0.9\linewidth]{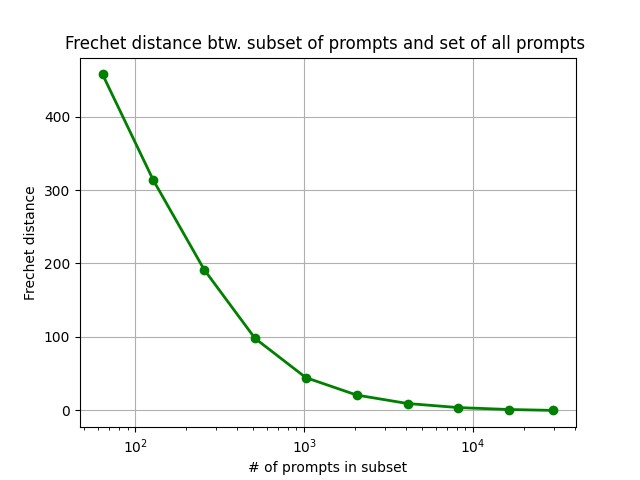}
    \caption{Fr\'echet distance between MS-COCO-$30$K valid set prompts and its subsets of various sizes. As can be seen, a subset of $5$K randomly sampled prompts adequately represents MS-COCO-$30$K's diversity, whereas smaller subsets, e.g., $150$ samples, show a large Fr\'echet distance and therefore are not good representations.}
    \label{fig:mscoco_prompts}
\end{figure}

Next we present our sampling methodology. We evaluate the diversity of a random subset of prompts by measuring its Fr\'echet distance to the entire MS-COCO-$30$K validation set prompts. Note that computing this measure does not use any images, as it is the Fr\'echet distance between two distributions of prompts' CLIP text encodings. As \cref{fig:mscoco_prompts} shows, a subset of $5$K randomly sampled prompts adequately represents MS-COCO-$30$K's diversity, whereas smaller subsets, such as those with $150$ samples, show a large Fr\'echet distance from the original set and therefore are not good representations. To further reduce the number of samples, we use the classifier-free guidance to generate images for the $5$K subset and rank them based on their CLIP score. Then, we use the $90$th, $50$th, and $10$th percentiles, representing the high-performing, middle-performing, and low-performing prompts, respectively, to form a final subset of $150$ prompts for our human evaluations. These two steps allow us to substantially reduce the resources needed for human evaluation while ensuring that the prompt subset is both diverse and fair. For example with $17$ human evaluators each rating $30$ pairs of images, we obtain an average of $1.7$ evaluations per each of the $300$ independent pairs from two tests.

\begin{figure}
  \centering
    \includegraphics[width=0.9\linewidth]{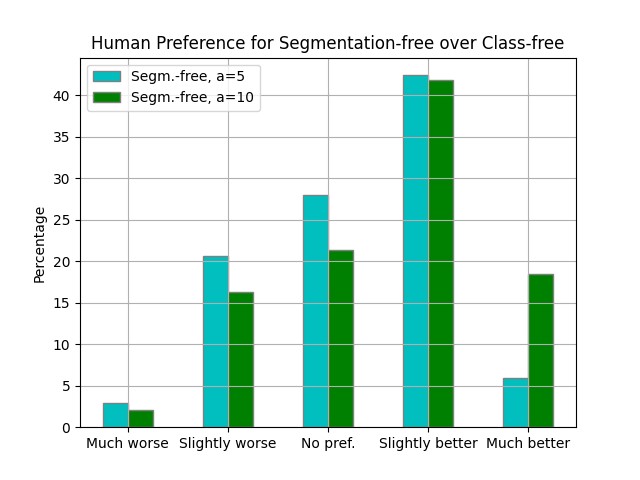}
    \caption{Preferences of human evaluators comparing segmentation-free to classifier-free guidance. Human evaluators preferred segmentation-free guidance $a = 10$ over classifier-free $60\%$ to $19\%$, with $18\%$ of occasions showing a strong preference.}
    \label{fig:subjective_hists}
\end{figure}

\cref{fig:subjective_hists} shows the subjective results of comparing segmentation-free to classifier-free guidance. Human evaluators favored segmentation-free guidance in $60\%$ of cases (with a segmentation-free scale of 10), compared to $19\%$ for classifier-free guidance. More significantly, human evaluators indicated a strong preference for segmentation-free guidance in $18\%$ of cases, compared to $2\%$ for classifier-free guidance. For visual comparisons, c.f., \cref{fig:comp}, where images for few user specified prompts have been generated.

%% file: sec/5_conclusion.tex
\section{Conclusion}
\label{sec:conclusion}

We presented segmentation-free guidance, a novel guidance method for text-to-image diffusion models like Stable Diffusion. Our method does not increase the computational load. It does not require retraining or fine-tuning. Segmentation-free guidance uses the diffusion model as an implied segmentation network to dynamically customize the negative prompt for each image patch, by focusing on the most relevant concept from the prompt. We evaluated segmentation-free guidance both objectively, using FID, CLIP, IS, and PickScore, and subjectively, through human evaluators. For the subjective evaluation, we proposed a methodology for reducing the number of prompts in a dataset like MS-COCO-30K to keep the number of human evaluations manageable while ensuring that the selected subset is both representative in terms of diversity and fair in terms of model performance. The results showed the superiority of our segmentation-free guidance to the widely used classifier-free method. Human evaluators preferred segmentation-free guidance over classifier-free $60\%$ to $19\%$, with $18\%$ of occasions showing a strong preference. PickScore win-rate, a recently proposed metric mimicking human preference, indicated a preference for our method over classifier-free, too.